\documentclass[10pt, a4paper]{article}

\usepackage[]{lrec-coling2024} 
\usepackage{inconsolata}
\usepackage{graphicx}
\usepackage{booktabs}
\usepackage{subfig}
\usepackage{tabularx}
\usepackage{color}
\usepackage{amsmath}
\usepackage{amssymb}
\usepackage{enumitem}

\title{Enhancing Semantics in Multimodal Chain of Thought via Soft Negative Sampling}

\name{Guangmin Zheng, Jin Wang{*}, Xiaobing Zhou, Xuejie Zhang \thanks{*Corresponding author}} 

\address{School of Information Science and Engineering \\
        Yunnan University \\
         Kunming, China \\
         gmzheng@mail.ynu.edu.cn, \{wangjin, zhouxb, xjzhang\}@ynu.edu.cn\\}

\abstract{
Chain of thought (CoT) has proven useful for problems requiring complex reasoning. Many of these problems are both textual and multimodal. Given the inputs in different modalities, a model generates a rationale and then uses it to answer a question. Because of the hallucination issue, the generated soft negative rationales with high textual quality but illogical semantics do not always help improve answer accuracy. This study proposes a rationale generation method using soft negative sampling (SNSE-CoT) to mitigate hallucinations in multimodal CoT. Five methods were applied to generate soft negative samples that shared highly similar text but had different semantics from the original. Bidirectional margin loss (BML) was applied to introduce them into the traditional contrastive learning framework that involves only positive and negative samples. Extensive experiments on the ScienceQA dataset demonstrated the effectiveness of the proposed method. Code and data are released at \url{https://github.com/zgMin/SNSE-CoT}.
 \\ \newline  }

\begin{document}

\maketitleabstract

\section{Introduction}
\label{sec1}
\begin{figure*}[t!]
    \centering  
    \centerline{\includegraphics[width=1\textwidth]{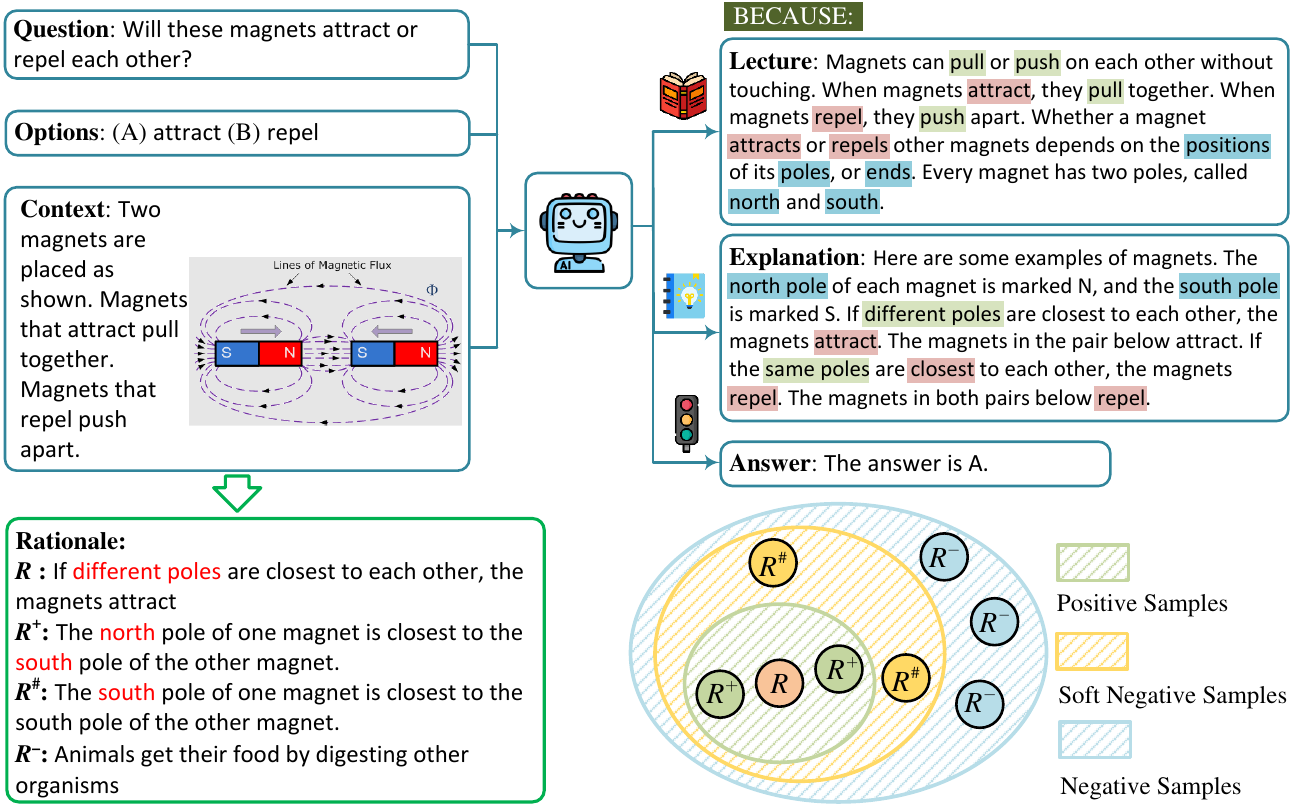}}
    \caption{The latent distribution of the samples. $R^+$ represents positive samples, $R^\#$ represents soft negative samples, and $R^-$ represents negative samples.}
    \label{a1}
\end{figure*}

Artificial intelligent systems have long been aimed at behaving dependably and learning complicated tasks quickly. As humans, we can use an explicit chain-of-thought (CoT) reasoning process, which is often articulated as an explanation, to make dependable decisions \cite{Wei2022}. Nevertheless, for a specific task, machine learning models are often trained using a large number of input--output samples. These black-box approaches only produce a final decision without consistently disclosing the underlying reasoning. CoT methods have recently been demonstrated to be extremely useful for large-scale language models (LLMs) in handling tasks that require complex reasoning. Most previous studies have focused only on language modalities, whereas inference may exist in multiple modalities, such as visual question answering (VQA). Given the inputs in different modalities, an intelligent system is required to infer answers using multi-hop intermediate reasoning.

Consider the ideas that an individual may have in response to the inquiry shown in Figure \ref{a1}. An individual can start by remembering the information about the definition of a magnetic force learned from textbooks as a lecture: \textit{"… Whether a magnet attracts or repels other magnets depends on the positions of its poles, or ends … If different poles are closest to each other, the magnets attract … If the same poles are closest to each other, the magnets repel …"}. Then, a chain of reasoning can be formed as an explanation: \textit{"The north pole of one magnet is closest to the south pole of the other magnet. $\to$ Poles that are different attract. $\to$ These magnets will attract each other."}. This finally leads to the correct answer: \textit{"These magnets attract each other."}. 

Early exploration of multimodal CoT involved transforming the inputs of different modalities into one modality and prompting LLMs to answer. One viable solution is to extract the caption of an image and concatenate it using the original language modality as input \cite{Lu2022,Lu2023}. However, a simplified caption cannot encompass all the details expressed by the image, leading to information loss in the reasoning process. Using only these captions may result in a lack of mutual synergy in the latent space of the multimodality. 
An alternative solution to facilitate the interaction between modalities is to fine-tune small models with cross-attentions to align multimodal features. Nevertheless, previous studies have shown that models trained with fewer than 100 billion parameters tend to produce illogical CoTs with hallucinated rationales \cite{Ho2022,Magister2022}. The challenge lies in the fact that the language model (LM) does not see images during pretraining and thus has no information about visual elements or methods to exploit vision features.

Recent studies have suggested incorporating both language and vision modalities into a two-stage framework, that is, rational generation and answer inference \cite{ZZhang2023}. Instead of prompting the LM with an image caption, the vision features are extracted using a vision encoder and fed to a decoder along with the encoded language representation. 

Although vision features are beneficial for better rationale generation, many reported errors stem from hallucinations \cite{ZZhang2023}. Considering the same example in Figure \ref{a1}, an appropriate rationale is, \textit{The \underline{north} pole of one magnet is closest to the south of the other magnet}. However, simply modifying one word can make the rationale unreasonable, that is, \textit{The \underline{south} pole of one magnet is closest to the south of the other magnet}. For the decoder of generation, this inappropriate rationale can achieve an extremely low negative log-likelihood but will finally mislead the answer inference.

One viable solution to mitigate the hallucinated generation is to treat these inappropriate rationales as negative samples for contrastive learning. Negative samples are difficult to exclude if textual features are used to measure the distribution of the latent space. Several recent studies have defined these rationales as soft negative samples. Here, \textit{negative} denotes that the samples differ semantically from the originals, whereas \textit{soft} denotes that the samples share many textual similarities with the originals and cannot be simply regarded as pure negative samples.

This study proposes a rationale generation method using soft negative sampling (SNSE-CoT) to mitigate hallucinations in multimodal CoT. Contrastive learning was introduced to enhance rationale generation, and five methods were applied to generate soft negative samples that shared highly similar text but had different semantics from the original. Bidirectional margin loss (BML) was applied to introduce them into the traditional contrastive learning framework that involves only positive and negative samples.

Extensive experiments were conducted on the ScienceQA dataset \cite{Lu2022} to validate the effectiveness of the SNSE-CoT. The results showed that the proposed model outperformed models of previous studies in most categories for multimodal CoT.

The remainder of this paper is organized as follows. Section \ref{sec2} reviews the preliminary knowledge. Section \ref{sec3} describes the proposed SNSE-CoT in detail. Section \ref{sec4} summarizes the experimental settings and empirical results. Section \ref{sec5} briefly reviews the related works. Conclusions are drawn in Section \ref{sec6}.

\section{Preliminary}
\label{sec2}
\begin{figure}[t!]
\centerline{\includegraphics[width=0.48\textwidth]{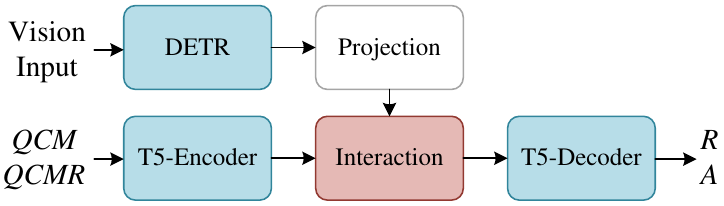}}
\caption{The overall architecture of the two-stage model.}
\label{a2}
\end{figure}
A multimodal CoT applies two-stage training, including rationale generation and answer inference. Figure \ref{a2} shows the overall architecture of the two-stage model. An input $X$ consists of an image input ${{X}_{v}}$ and a language input ${{X}_{l}}$, where $Q$ denotes the question text and $C$ denotes the context text. The goal is to select correct answer $A$ from multiple options $M$. To implement a multimodal CoT, the first stage is rationale generation, in which the model is required to generate a rationale.
\begin{align}
R=f(X)
\label{e1}
\end{align}

For answer inference, rationale $R$ is appended to original language input ${{\bar{X}}_{l}}$ as a new input, that is, ${{\hat{X}}_{l}}={{\bar{X}}_{l}}\circ R$, where $\circ$ is a concatenate operator. Then, updated input ${X}'=\{{{\hat{X}}_{l}},{{X}_{v}}\}$ is fed into the model to infer the final answer.
\begin{align}
A=g({X}')
\label{e2}
\end{align}

In both stages, two independent transformer-based models $f$ and $g$ with the same architecture are trained. The aliases of both the stages can be represented as $X\to R$ and $X'\to A$, respectively. 


\noindent \textbf{Encoding.} For both stages, the vision and language inputs are ${{X}_{v}}$ and ${{X}_{l}}\in \{{{\bar{X}}_{l}},{{\hat{X}}_{l}}\}$, respectively, where ${{\bar{X}}_{l}}$ is used for rationale generation, and ${{\hat{X}}_{l}}$ is used for answer inference. The T5 encoder \cite{Carion2020} is used to encode language input ${{\hat{X}}_{l}}$, and the DETR \cite{Raffel2020} vectorizes vision input ${{X}_{v}}$ into vision features.
\begin{align}
{{H}_{l}}&=\text{T5-Encoder}({{X}_{l}})\label{e3} \\
{{H}_{v}}&={{W}_{h}}\cdot \text{DETR}({{X}_{v}})
\label{e4}
\end{align}

\noindent where ${{H}_{l}}\in {{\mathbb{R}}^{n\times d}}$ is the hidden representation of the last layer of the T5 encoder, $n$ denotes the length of the language input, and $d$ denotes the dimensionality. ${{H}_{v}}\in {{\mathbb{R}}^{m\times d}}$ is a vision feature, where $m$ denotes the number of image patches, and ${{W}_{h}}\in {{\mathbb{R}}^{d\times {{d}_{v}}}}$ is a linear projection used to transform the dimensionality from $d_v$ to $d$. 

\noindent \textbf{Interaction.} To integrate both vision and language representations, cross-attention with single-head self-attention is used to align text tokens with image tokens. 
\begin{align}
{{\hat{H}}_{v}}=\text{softmax}\!\left[ \frac{({{W}_{Q}}{{H}_{l}})\cdot {{({{W}_{K}}{{H}_{v}})}^{\top }}}{\sqrt{d}} \right]\!({{W}_{V}}{{H}_{v}})
\label{e5}
\end{align}

\noindent where $W_Q$, $W_K$, and $W_V$ denote weight matrices for self-attention. Subsequently, a gated fusion mechanism is applied to integrate both features that is denoted as follows:
\begin{align}
&\sigma =\text{sigmoid}({{W}_{l}}{{H}_{l}}+{{W}_{v}}{{\hat{H}}_{v}})
\label{e6} \\
&{{H}_{\text{Enc }}}=(1-\sigma )\cdot {{H}_{l}}+\sigma \cdot {{\hat{H}}_{v}}
\label{e7}
\end{align}

\noindent where ${{W}_{l}}$ and ${{W}_{v}}$ denote trainable matrices.

\noindent \textbf{Decoding.} The model predicts the probability of generating a target $Y\in \{R,A\}$ with length $N$. The models are trained by minimizing the negative likelihood loss. 
\begin{align}
{{\mathcal{L}}_{\text{NLL}}}=-\sum\limits_{i=1}^{N}{\log {{p}_{\theta }}({{Y}_{i}}|{{X}_{l}},{{X}_{v}},{{Y}_{<i}})}
\label{e8}
\end{align}

\noindent where $\theta $ denotes all trainable parameters of either $f$ or $g$.

\section{Mitigating Hallucinated Generation}
\label{sec3}
Both the quality and semantic correctness of rationale generation ultimately affect the choice of answer inference. This study proposes enhancing the ability to discriminate soft negative samples by introducing a bidirectional margin loss.
\subsection{Soft Negative Sampling}
A high-quality soft negative sample is indistinguishable from the target sample and has different semantics for key information. Typically, high-quality soft negative samples are difficult to generate but can be obtained by modifying the target sample and observing the following three principles.

\begin{figure*}[t!]    
  \centering            
  \subfloat[Principal 1]   
  {
        \label{a}
        \includegraphics[width=0.265\textwidth]{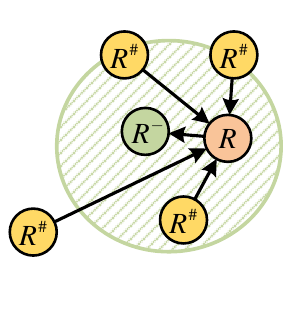}
  }
  \subfloat[Principal 2]
  {
      \label{b}
        \includegraphics[width=0.33\textwidth]{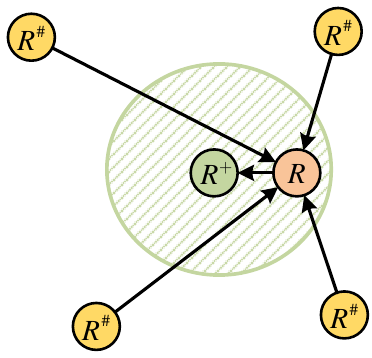}
  }
  \subfloat[Principal 3]
  {
      \label{c}
        \includegraphics[width=0.33\textwidth]{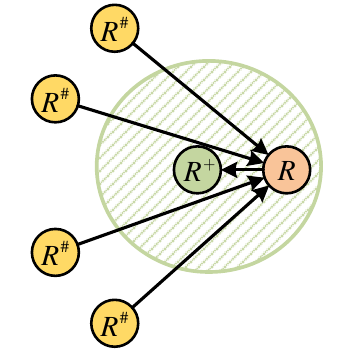}
  }
  \caption{Possible situations arising from non-observance of the modification principle. $R$ is the generated rationale.}    
  \label{a3}            
\end{figure*}

\begin{itemize}[itemsep=2pt,topsep=0pt,parsep=0pt]
    \item \textbf{Principal 1:} \textit{Soft negative samples have different semantics from the target sample for key information.} As shown in Figure \ref{a3}\subref{a}, if this cannot be guaranteed, then the introduction of soft negative samples may move the generated samples away from the positive samples.
    \item \textbf{Principal 2:} \textit{The modification of the target sample should be few.} Soft negative samples have high text similarity and a highly similar distribution with the target sample. The fewer the modifications to the target sample, the higher the text similarity and the higher the difficulty to distinguish the soft negative samples from the target sample. The number of modifications can be moderately increased according to the increase in the length of the target sample. As shown in Figure \ref{a3}\subref{b}, if excessive modifications are introduced, the soft negative samples are far from the distribution of the positive samples, resulting in a negligible impact.
    \item \textbf{Principal 3:} \textit{The generation of soft negative samples should be multiple and random.} Multiple soft negative samples should be distributed as evenly as possible around the positive sample region in the latent space. As shown in Figure \ref{a3}\subref{c}, when the soft negative samples are on the same side as the target sample, they may push the generated samples away from the positive sample region instead of toward the target sample.
\end{itemize}

The proposed soft negative sampling is specified using five methods as follows:
\begin{itemize}[itemsep=2pt,topsep=0pt,parsep=0pt]
    \item \textbf{Affirmation-Negation Transformation.} Apply explicit negation with negative words, and based on the parsing information of SpaCy\footnote{\url{https://github.com/explosion/spaCy.}}, convert the sentences into syntactically correct and semantically clear negations. 
    \item \textbf{Number Transformation.} Randomly select some numbers in the sentence and replace them with random numbers of equal length. When the length is greater than 1, ensure that the first number is not 0.
    \item \textbf{Orientation Transformation.} Randomly select some orientation words in the sentence and replace them with the opposite direction.
    \item \textbf{Unit Transformation.} Randomly select some unit words in the sentence and replace them with other units of the same category at random.
    \item \textbf{Option Transformation.} Randomly select some strings that contain the correct option in the sentence and replace them with other options. 
\end{itemize}

Specifically, if other transformations fail to be implemented, we use an affirmation-negation transformation.

The modification of key information varies depending on the form of the sample. For multimodal reasoning, the explanation is modified first and then the\ lecture. The explanation provides reasoning ideas for a specific problem, which is considered to contain more important information, whereas the lecture typically provides methodological guidance, examples, and background knowledge for solving the problem. The rationale is a concatenation of lectures and explanations. If the rationale is devoid of any content , we use \textit{Not} as a rationale.

\subsection{Bidirectional Margin Loss}
The generated positive and soft negative samples of the target rationale are further extracted using the following steps:
\begin{align}
&E_R = \operatorname{emb}(R)
\label{e9} \\
&h=\operatorname{mean}({{W}_{R}}{{E}_{R}}+{{b}_{R}})
\label{e10}
\end{align}

\noindent where $\text{emb}(\cdot)$ represents the embedding layer of the encoder in the rationale generation, and $W_R$ and $b_R$ represent the trainable matrix and bias, respectively.

Cosine similarity difference $\Delta$ between positive and soft negative pairs is calculated as follows:
\begin{align}
\Delta =\text{cos}({{h}_{i}},h_{ij}^{\#})-\text{cos}({{h}_{i}},h_{i}^{+})
\label{e11}
\end{align}

\noindent where $\text{cos}(\cdot )$ is the cosine similarity, $h_i$ is the representation of the generated sample, $h_{i}^{+}$ denotes the corresponding positive sample, and $h_{ij}^{\#}$ denotes the corresponding $j$-th soft negative sample. BML is used to model semantic similarity differences.
\begin{align}
{{\mathcal{L}}_{\text{BML}}}= \frac{1}{k}\sum\limits_{j=1}^{k}{(\operatorname{ReLU}(}\Delta \!+\!\alpha )+\operatorname{ReLU}(\!-\!\Delta \!-\!\beta ))
\label{e12}
\end{align}

\noindent where $k$ denotes the number of soft negative samples corresponding to each target sample, and $\alpha$ and $\beta$ denote the bottom and upper  differences in semantic similarity difference between the positive and soft negative pairs, respectively. The BML aims to constrain $\Delta$ within an interval of $\Delta \in [-\beta ,-\alpha ]$.

\subsection{Training Objective}
The training objectives for rationale generation and answer inference are respectively expressed as follows:
\begin{align}
&{{\mathcal{L}}_{\text{RG}}}={{\mathcal{L}}_{\text{NLL}}}+\lambda {{\mathcal{L}}_{\text{BML}}}
\label{e13} \\
&{{\mathcal{L}}_{\text{AI}}}={{\mathcal{L}}_{\text{NLL}}}
\label{e14}
\end{align}

\noindent where $\lambda$ is used to balance the two losses.

\section{Experiments}
\label{sec4}
\subsection{Dataset}
Empirical experiments were conducted using the ScienceQA benchmark \cite{Lu2022}, which is the first multimodal question answering dataset with a detailed CoT. Science QA features 26 topics, 127 categories, and 379 skills, covering a wide range of domains. The benchmark dataset was divided into training, validation, and test splits using 12,726, 4,241, and 4,241 examples, respectively. Eight question categories included were natural science, social science, language science, textual context, pictorial context, no context, grades 1--6, and grades 7--12. For the rationale generation stage, the optimal model was selected based on the ROUGE-L score. The predicted results were evaluated based on accuracy.
\subsection{Implementation Details}
UnifiedQA \cite{Khashabi2020} was used to initialize the T5 model in two stages because it achieved the best fine-tuning results in the experiments by \citet{Lu2022}. The model was fine-tuned for up to 20 epochs at a learning rate of 5e-5. The maximum input sequence lengths were 512 and 64 in the rationale generation and answer inference stages, respectively. For the soft negative samples, one sample was generated for each generation method in each round, and each sample was modified randomly in only one place. $\alpha$, $\beta$ and $\lambda$ were set to 0.1, 0.3 and 0.1, respectively. These three parameters are further explored in Section \ref{Hyperparameter}. The random seed number was set to 42 to ensure reproducibility.
\subsection{Baselines}
For comparison, three categories of baseline models were selected as follows:
\begin{itemize}[itemsep=2pt,topsep=0pt,parsep=0pt]
    \item MCAN \cite{Yu2019}, Top-Down \cite{Anderson2018}, BAN \cite{Kim2018}, DFAF \cite{Gao2019}, ViLT \cite{Kim2021}, Patch-TRM \cite{Lu2021}, and VisualBERT \cite{Li2020}.
    \item UnifiedQA$_{\mathrm{Base}}$ w/CoT \cite{Lu2022}, Multimodal-CoT$_{\mathrm{Base}}$, and Multimodal-CoT$_{\mathrm{Large}}$ \cite{ZZhang2023}.
    \item GPT-3.5 w/CoT \cite{Lu2022}, LLaMA-Adapter \cite{Zhang2023}, LLaVa, LLaVa (GPT-4) \cite{Liu2023}, and Chameleon (GPT-4) \cite{Lu2023}.
\end{itemize}

More details are presented in Appendix \ref{A}.

\subsection{Comparative Results}
\begin{table*}[t!]
\centering
\resizebox{\linewidth}{!}{
\begin{tabular}{c|c|cccccccc|c}
\toprule
Model               & Size & NAT            & SOC            & LAN            & TXT            & IMG            & NO             & G1-6           & G7-12          & Avg            \\ \hline
Human               & -    & 90.23          & 84.97          & 87.48          & 89.60          & 87.50          & 88.10          & 91.59          & 82.42          & 88.40          \\ \hline
MCAN                & 95M  & 56.08          & 46.23          & 58.09          & 59.43          & 51.17          & 55.40          & 51.65          & 59.72          & 54.54          \\
Top-Down            & 70M  & 59.50          & 54.33          & 61.82          & 62.90          & 54.88          & 59.79          & 57.27          & 62.16          & 59.02          \\
BAN                 & 112M & 60.88          & 46.57          & 66.64          & 62.61          & 52.60          & 65.51          & 56.83          & 63.94          & 59.37          \\
DFAF                & 74M  & 64.03          & 48.82          & 63.55          & 65.88          & 54.49          & 64.11          & 57.12          & 67.17          & 60.72          \\
ViLT                & 113M & 60.48          & 63.89          & 60.27          & 63.20          & 61.38          & 57.00          & 60.72          & 61.90          & 61.14          \\
Patch-TRM           & 90M  & 65.19          & 46.79          & 65.55          & 66.96          & 55.28          & 64.95          & 58.04          & 67.50          & 61.42          \\
VisualBERT          & 111M & 59.33          & 69.18          & 61.18          & 62.71          & 62.17          & 58.54          & 62.96          & 59.92          & 61.87          \\ \hline
UnifiedQA$_{\mathrm{Base}}$ w/CoT & 223M & 71.00          & 76.04          & 78.91          & 66.42          & 66.53          & 81.81          & 77.06          & 68.82          & 74.11          \\
GPT-3.5 w/CoT       & 175B & 75.44          & 70.87          & 78.09          & 74.68          & 67.43          & 79.93          & 78.23          & 69.68          & 75.17          \\
Multimodal-CoT$_{\mathrm{Base}}$  & 223M & 87.52          & 77.17          & 85.82          & 87.88          & 82.90          & 86.83          & 84.65          & 85.37          & 84.91          \\
Multimodal-CoT$_{\mathrm{Large}}$ & 738M & 95.91          & 82.00          & 90.82          & 95.26          & 88.80          & 92.89          & 92.44          & 90.31          & 91.68          \\ \hline
LLaMA-Adapter       & 6B   & 84.37          & 88.30          & 84.36          & 83.72          & 80.32          & 86.90          & 85.83          & 84.05          & 85.19          \\
LLaVa               & 13B  & 90.36          & 95.95          & 88.00          & 89.49          & 88.00          & 90.66          & 90.93          & 90.90          & 90.92          \\
LLaVa (GPT-4)       & 13B  & 91.56          & \textbf{96.74} & 91.09          & 90.62          & 88.99          & 93.52          & 92.73          & 92.16          & 92.53          \\
Chameleon (GPT-4)   & -    & 89.83          & 74.13          & 89.82          & 88.27          & 77.64          & 92.13          & 88.03          & 83.72          & 86.54          \\ \hline
SNSE-CoT$_{\mathrm{Base}}$        & 223M & 90.05          & 78.85          & 89.09          & 89.64          & 84.78          & 90.38          & 87.67          & 87.08          & 87.46          \\
SNSE-CoT$_{\mathrm{Large}}$       & 738M & \textbf{96.80} & 90.33          & \textbf{93.09} & \textbf{96.73} & \textbf{93.36} & \textbf{94.08} & \textbf{94.71} & \textbf{94.07} & \textbf{94.48} \\ \bottomrule
\end{tabular}
}
\caption{Comparison of the experiment results (\%). Size = backbone model size. Question classes: NAT = natural science, SOC = social science, LAN = language science, TXT = text context, IMG = image context, NO = no context, G1-6 = grades 1-6, G7-12 = grades 7-12. Part 1: Human performance; Part 2: Traditional VQA; Part 3: Small model with CoT; Part 4: Large model with CoT; Part 5: Our SNSE-CoT results. Results in bold are the best performance.}
\label{result}
\end{table*}
Table \ref{result} summarizes the experimental results of the proposed method relative to those of the baselines. SNSE-CoT$_{\mathrm{Base}}$ performed similarly to humans. SNSE-CoT$_{\mathrm{Large}}$ outperformed all the previous methods to achieve the current best performance. Compared with Multimodal-CoT, the average performance of SNSE-CoT increased by approximately 2.5 to 3\%. Moreover, SNSE-CoT$_{\mathrm{Large}}$ was worse than LLaVa (GPT-4) only for social science problems (SOC) for all types of problems, indicating that the soft negative sample generation method designed in this study improved the correctness of the model for generating various types of CoTs. In particular, SNSE-CoT$_{\mathrm{Large}}$ achieved substantial improvement in problems with paired images (IMG), becoming the first method to exceed 90\% performance on this type of problem. This indicated that the contrastive learning approach strengthened the model’s ability to accurately understand images of specific problems. 

\subsection{Hyperparameter Fine-Tuning}
\label{Hyperparameter}

\begin{figure}[t!]
\centerline{\includegraphics[width=0.48\textwidth]{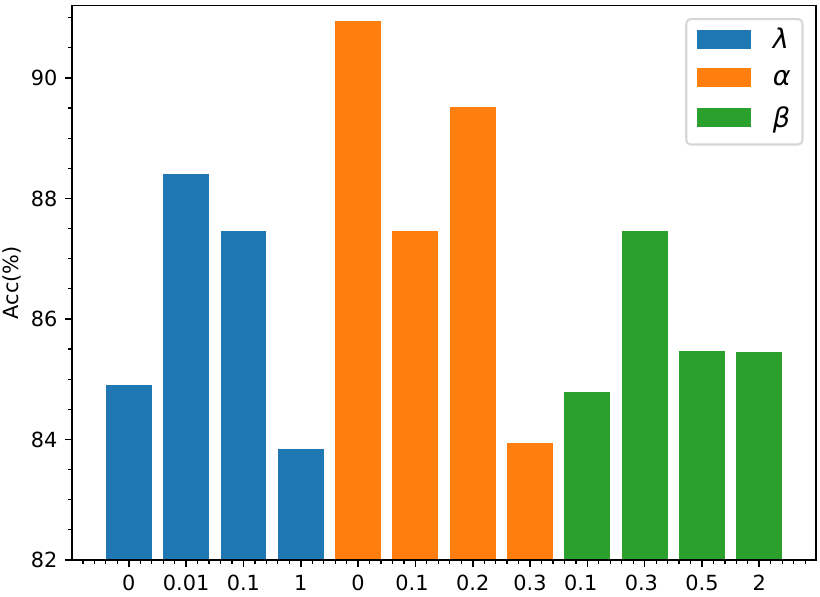}}
\caption{Hyperparameters fine-tuning.}
\label{hyper}
\end{figure}

To explore the effect of hyperparameters $\lambda$ in Eq. \eqref{e13} and $\alpha$ and $\beta$ in Eq. \eqref{e12}, a grid strategy was used. For balanced parameter $\lambda$, candidate set $\{0,\ 0.01,\ 0.1,\ 1\}$ was used; for bottom difference $\alpha$ of the cosine similarity difference, the candidate set $\{0,\ 0.1,\ 0.2,\ 0.3\}$ was used; and for upper difference $\beta$ of the cosine similarity difference, candidate set $\{0.1,\ 0.3,\ 0.5,\ 2\}$ was used. $\beta=\ 2$ is a special value indicating that the upper difference is ignored because the range of the cosine similarity is -1.0 to 1.0. The results of the parametric analysis are shown in Figure \ref{hyper}.

Balance parameter $\lambda$ affects the strength of the soft negative sample rejection. If $\lambda$ is extremely small, soft negative samples play a limited role and performance grows less. When $\lambda$ is extremely large, the generated samples have difficulty in clustering toward the target center, and the performance may even become worse. $\alpha$ and $\beta$ constrain the range of cosine similarity difference, and an appropriate bottom and upper difference can play a better role in soft negative samples. The effect of bottom difference $\alpha$ on performance is more important.

\subsection{Ablation Studies}

Table \ref{ablation} reports the results of the ablation experiments. To investigate the effectiveness of the proposed soft negative sample generation method, the number, orientation, unit, and option transformations were removed separately. For the number transformation, a slight decrease in performance was observed after removal, indicating that the method did not considerably enhance the model’s numerical understanding. For the orientation and unit transformations, which further enhance the model’s understanding of the map and its ability to compare values, a decrease in performance of approximately 1\% was observed after removal. For the option transformation, a significant decrease in model performance was observed after removal because it is a general transformation that can modify keywords well for QA problems. 

The principle of soft negative sample generation was also analyzed. The randomness of \textbf{Principle 3} was ablated by fixing the soft negative samples for each epoch. As observed from the results, the ablation of randomness did not guarantee that the soft negative samples were evenly distributed around the positive sample area, and the performance yielded a significant degradation of approximately 2\%.

\begin{table}[t!]
\centering
\begin{tabular}{lc}
\toprule
\multicolumn{1}{c}{Model}                          & Avg   \\ \midrule
SNSE-CoT$_{\mathrm{Base}}$       & 87.46 \\
\,\,\,\,\,\,\,\,w/o number                       & 87.12 \\
\,\,\,\,\,\,\,\,w/o orientation                  & 86.23 \\
\,\,\,\,\,\,\,\,w/o unit                         & 86.03 \\
\,\,\,\,\,\,\,\,w/o option                       & 85.69 \\
\,\,\,\,\,\,\,\,w/o random                       & 85.57 \\ \bottomrule
\end{tabular}
\caption{Ablation study on SNSE-CoT (\%).}
\label{ablation}
\end{table}

\begin{table}[t!]
\centering
\begin{tabular}{ccccc}
\toprule
Number  & 1 & 2     & 3     & all   \\ \midrule
Changes & 0 & -0.59 & +0.08 & +0.17 \\ \bottomrule
\end{tabular}
\caption{The impact of the amount of modification on performance (\%). All means all modifications. }
\label{mod_num}
\end{table}

To explore the effect of the number of modifications mentioned in \textbf{Principle 2}, the relevant experimental results are reported in Table \ref{mod_num}. The small effect of the number of modifications on the performance might be due to the small number of modifiable positions in each CoT and the limited increase in the distance of the soft negative samples from the target center as the number of modifications increased.

\subsection{Visual Latent Distribution of Samples}

\begin{figure}[t!]
\centerline{\includegraphics[width=0.49\textwidth]{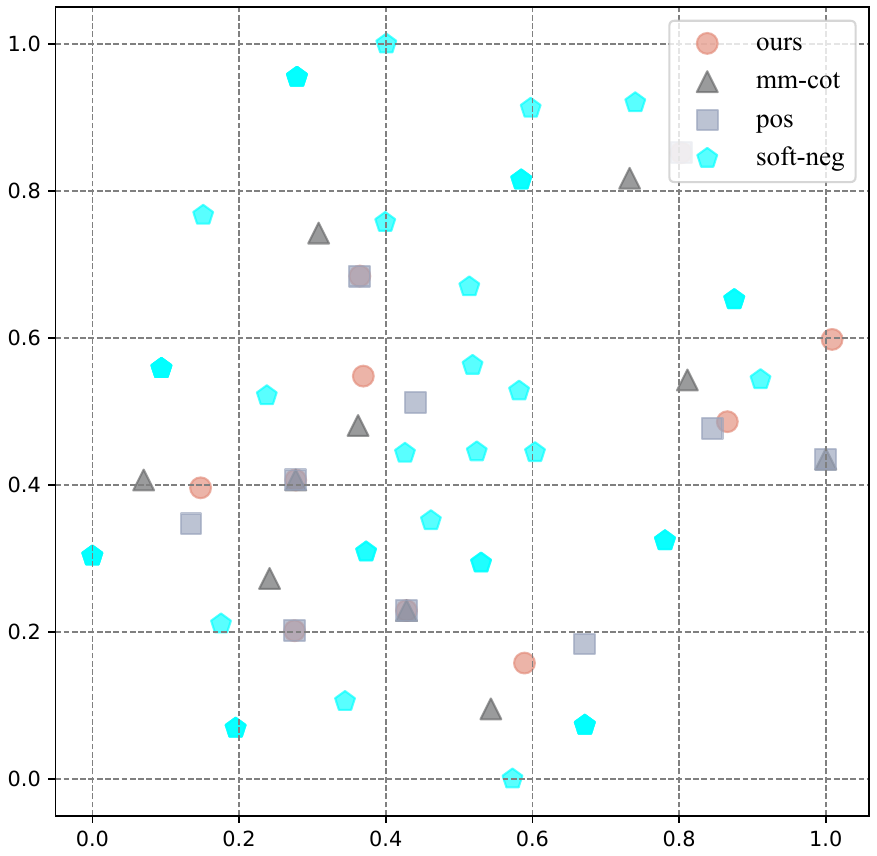}}
\caption{Visual latent distribution. “ours” represents samples generated by SNSE-CoT, “mm-cot” represents samples generated by Multimodal-CoT, “pos” represents positive samples, and “soft-neg” represents soft negative samples.}
\label{visual}
\end{figure}

We randomly selected ten samples, and the samples generated by SNSE-CoT, samples generated by Multimodal-CoT, positive samples, and soft negative samples of each sample were further extracted as feature $h$ using Eq. \eqref{e9} and \eqref{e10}: These functions were visualized using the tSNE tool.

Figure \ref{visual} illustrates this visualization. The soft negative samples were largely evenly distributed around the positive samples. This was consistent with the expectations of this study, and demonstrated the high quality of the soft negative samples generated by the proposed method. However, soft negative samples were still distributed on one side or far from each other, and our method must be improved. In terms of the distribution distance between the samples generated by Multimodal-CoT, the samples generated by SNSE-CoT, and the target positive samples, the distribution of the samples generated by SNSE-CoT was generally closer to the target samples, or even overlapped. However, worse cases existed, in which the samples generated by SNSE-CoT were pushed far away from the positive sample area by the soft negative samples, suggesting that the generation of soft negative samples should more carefully follow the generation principles proposed in this study.

\subsection{Case Analysis}
\begin{table*}[t!]
\centering
\begin{tabularx}{\textwidth} {>{\hsize=0.07\hsize}X
>{\hsize=.49\hsize}X
>{\hsize=.52\hsize}X
>{\hsize=.49\hsize}X}
\toprule
ID    & \multicolumn{1}{c}{Golden CoT}                                                                                               & \multicolumn{1}{c}{MultiModal-CoT}                                                                                            & \multicolumn{1}{c}{SNSE-CoT}                                                                                                 \\ \midrule
6902  & … This is the \textcolor{green}{Pacific Ocean}.                                                                                                 & … This is the \textcolor{red}{Indian Ocean}.                                                                                                   & … This is the \textcolor{green}{Pacific Ocean}.                                                                                                \\
7301  & … there were \textcolor{green}{6} solute particles on the left side of the membrane and \textcolor{green}{2} solute particles on … & … there were \textcolor{red}{5} solute particles on the left side of the membrane and \textcolor{red}{6} solute particles on … & … there were \textcolor{red}{5} solute particles on the left side of the membrane and \textcolor{green}{2} solute particles on … \\
10130 & … Animals \textcolor{green}{get} their food by digesting other organisms.                                                                       & … Animals \textcolor{red}{cannot} their food by digesting other organisms.                                                                     & … Animals \textcolor{green}{get} their food by digesting other organisms.                                                                     \\
12785 & … the temperature of the air on a hot day is \textcolor{green}{36°C}. 36°F is too \textcolor{green}{cold}.                                 & … the temperature of the air on a hot day is \textcolor{green}{36°F}. 36°F is too \textcolor{green}{hot}.                                   & … the temperature of the air on a hot day is \textcolor{green}{36°C}.n36°F is too \textcolor{green}{cold}..               \\ \bottomrule                
\end{tabularx}
\caption{CoT outputs of different models. ID indicates the identification number of the question in the ScienceQA dataset. Keywords are labeled with different colors, where red refers to wrong words, green refers to correct words. }
\label{case}
\end{table*}

Several examples were selected to further illustrate the effectiveness of SNSE-CoT in improving the quality of CoT generation. Table \ref{case} lists the selected examples corresponding to golden CoT, CoT generated by Multimodal-CoT, and CoT generated by SNSE-CoT. A more detailed case analysis is provided in Appendix \ref{C}. The results indicated that the proposed SNSE-CoT enhanced the map numerical aspects, affirmation-negation comprehension, and comparison of values in different units. In particular, the SNSE-CoT was still deficient in numerical comprehension and was only partially correct in the relevant example that corresponded to the results of the ablation experiment.

\section{Related Work}
\label{sec5}
\subsection{Traditional VQA}

VQA is a series of tasks that provides a picture and natural language question related to that picture, and the computer can produce the correct answer. Since the VQA task was first proposed \cite{Antol2015}, many VQA datasets \cite{Goyal2017,Hudson2019,Johnson2017} have been constructed to contribute to this research effort.

Researchers have proposed various approaches to improve the accuracy and interpretability of the models. Among them, joint embedding approaches \cite{Li2020} jointly encode the image and question and then decode the features of the mixed modality to generate the answer. More studies have focused on the application of attention mechanisms \cite{Anderson2018,Gao2019,Kim2018,yuan2023joint} that have shown that models are more effective at capturing key image parts based on questions. Compositional models \cite{Andreas2016,Xiong2016} provide a method to dynamically compose modules to generate answers based on the type of question. Other researchers have focused on introducing external knowledge bases \cite{Wu2016} to supplement the knowledge required to answer questions by retrieving knowledge bases. Each of these models has its own characteristics; however, they are all black-box models that output only answers.

ScienceQA datasets containing multimodal contexts and different topics in the scientific domain have been proposed. Most answers are annotated with lectures and explanations that allow multimodal CoT reasoning to be elicited and the reasoning process to be explicitly presented.

\subsection{CoT Reasoning}
CoT encourages LLMs to generate intermediate chains of reasoning to solve problems, and LLMs typically use two technical paradigms for CoT reasoning: zero-shot CoT \cite{Kojima2022} and few-shot CoT \cite{Wei2022,Zhang2022}. Few-shot CoT uses step-by-step reasoning demonstrations as the conditions for reasoning, each of which contains the question and chain of reasoning leading to the final answer and can be produced manually or automatically, called Manual-CoT \cite{Kojima2022} and Auto-CoT \cite{Zhang2022}, respectively. Effective demonstrations make few-shot CoT a stronger performer than zero-shot CoT and have attracted more research interest.

Certain studies have focused on the ability to inspire CoT reasoning in small models. However, models with 100 billion parameters tend to produce illogical CoTs, leading to incorrect answers \cite{Wei2022}. The performance substantially drops when small models are directly fine-tuned to generate CoTs to reason answers \cite{Lu2022}. Knowledge distillation is employed by fine-tuning the student model on the output of the CoT generated by the larger teacher model \cite{Ho2022,Magister2022}, and significant performance improvement is obtained.

As stated in Section \ref{sec1}, several studies focused on multimodal CoT reasoning. The critical challenge is to unify vision and language modalities. The images are converted into captions to prompt the LLMs for CoT inference \cite{Lu2022,Lu2023}. However, caption conversion loses considerable information, and researchers have attempted to interactively combine vision and language modalities in large and small models \cite{Liu2023,Zhang2023,ZZhang2023} to obtain remarkable results. For instance, \citet{ZZhang2023} indicated that the introduction of the visual modality supplemented more information and alleviated the hallucinations of the small model CoT, and the small model CoT reasoning ability made a qualitative leap.

\subsection{Contrastive Learning}
Contrastive learning allows the models to learn from both positive and negative samples. Three key issues in contrastive learning are construction of positive and negative samples, design of the encoder, and selection of the loss function. The development of contrastive learning can be divided into four phases.

In the first phase, the various methods and models are not unified, nor are the objective functions and agent tasks \cite{Wu2018,Ye2019}. In the second phase \cite{TChen2020,XChen2020}, the details tend to be uniform, objective function is InfoNCE or similar, and model is a combination of the encoder and projection head. Stronger data augmentation is used. Momentum encoders have been proposed to solve feature inconsistency problems. In the third phase, contrastive learning eliminates the use of negative samples \cite{XChen2020,Grill2020}; this stage is a summary generalization of all methods. In the fourth phase, most studies have focused on the use of contrastive learning in transformers \cite{Caron2021}.

Recently, the concept of soft negative samples \cite{Wang2022} was proposed to guide models to focus on semantic similarity and alleviate feature suppression.

Most negative samples for these methods originate from other samples within the same batch, and few studies have focused on the manner in which negative samples are generated.

\section{Conclusions}
\label{sec6}
This study proposed mitigating hallucinated rationale generation by using soft negative sampling for multimodal scientific quiz questions to generate more accurate CoTs. Specifically, we designed five high-quality soft negative sample generation methods: affirmation-negation transformation, number transformation, orientation transformation, unit transformation, and option transformation. Bidirectional margin loss was used to enable the model to distinguish between soft negative samples. Experimental results showed that the proposed method outperformed the methods in the previous studies on the ScienceQA benchmark dataset and validated the effectiveness of the proposed methods.

Future work will attempt to design a general method for automatically generating soft negative examples such that the model self-corrects for small semantic differences.




\section*{Acknowledgements}

This work was supported by the National Natural Science Foundation of China (NSFC) under Grant Nos.61966038 and 62266051, and the Exam-Exempted Postgraduate Research and Innovation Foundation of Yunnan University under Grant No.TM-23236806. The authors would like to thank the anonymous reviewers for their constructive comments.

\nocite{*}
\section*{Bibliographical References}\label{sec:reference}
\bibliographystyle{lrec-coling2024-natbib}
\bibliography{lrec-coling2024-example}

\label{lr:ref}
\bibliographystylelanguageresource{lrec-coling2024-natbib}
\bibliographylanguageresource{languageresource}

\appendix

\section{Baseline Models}
\label{A}
For comparison, three categories of baseline models are selected as follows: 

\subsection{Traditional VQA}
Traditional VQA baselines consider the question, context, and choices as the textual input and the image as the vision input, and predict the score distribution over choice candidates via a linear classifier.

\begin{itemize}[itemsep=2pt,topsep=0pt,parsep=0pt]
    \item MCAN \cite{Yu2019} designs self-attention units and guided attention units, and constructs a new deep co-attention network through the combination and stacking between them.
    \item Top-Down \cite{Anderson2018} proposes a new visual attention mechanism that combines bottom-up and top-down to allow attention to be calculated more naturally at the object and other prominent area level.
    \item BAN \cite{Kim2018} uses variants of multimodal residual networks for joint representation and finally classification by MLP to predict answers. 
    \item DFAF \cite{Gao2019} proposes a multimodal feature fusion method using external and internal modal information flow.
    \item ViLT \cite{Kim2021} and Patch-TRM \cite{Lu2021} parse the diagram in a pyramid layout and apply cross-modal transformers with attention mechanism to learn the meaningful joint diagram-question feature. 
    \item VisualBERT \cite{Li2020} is a visual language pretraining model using a self-attention mechanism to mine the relationship between text and image regions in the input image.
\end{itemize}

\subsection{Small Model with CoT}
\begin{itemize}[itemsep=2pt,topsep=0pt,parsep=0pt]
    \item UnifiedQA$_{\mathrm{Base}}$ w/CoT \cite{Lu2022} extracts the caption of an image and concatenates it with the original language modality as the input to fine-tune the LM.
    \item Multimodal-CoT$_{\mathrm{Base}}$ and Multimodal-CoT$_{\mathrm{Large}}$ \cite{ZZhang2023} use gated fusion mechanisms to combine language and visual modalities into a two-stage framework.
\end{itemize}

\begin{table*}[t!]
\centering
\begin{tabular}{l|cccccccc|c}
\toprule
\multicolumn{1}{c}{Model}              & NAT   & SOC   & LAN   & TXT   & IMG   & NO    & G1-6  & G7-12 & Avg   \\ \midrule
SNSE-CoT$_{\text{Base}}$        & 90.05 & 78.85 & 89.09 & 89.64 & 84.78 & 90.38 & 87.67 & 87.08 & 87.46 \\
\,\,\,\,\,\,\,\,w/o vision features & 75.63 & 70.76 & 79.82 & 74.18 & 68.51 & 83.15 & 75.83 & 73.66 & 75.41 \\
SNSE-CoT$_{\text{Large}}$        & 96.80 & 90.33 & 93.09 & 96.73 & 93.36 & 94.08 & 94.71 & 94.07 & 94.48 \\
\,\,\,\,\,\,\,\,w/o vision features & 82.09 & 82.34 & 89.42 & 88.60 & 81.35 & 90.10 & 84.69 & 83.31 & 85.84 \\ \bottomrule
\end{tabular}
\caption{Ablation results of vision features(\%).}
\label{B}
\end{table*}

\subsection{Large Model with CoT}
\begin{itemize}[itemsep=2pt,topsep=0pt,parsep=0pt]
    \item GPT-3.5 w/CoT \cite{Lu2022} uses the same method as that of UnifiedQA$_{\mathrm{Base}}$ w/CoT to prompt LLMs.
    \item LLaMA-Adapter \cite{Zhang2023} adjusts LLaMA \cite{Touvron2023} to an instruction-following model by fine-tuning the adapter to insert vision features into the LM. 
    \item LLaVa \cite{Liu2023} is a multimodal large model fine-tuned using a multimodal instruction dataset.
    \item LLaVa (GPT-4) \cite{Liu2023} collaborates with GPT-4 to first explain the cause and then infer the answer.
    \item Chameleon (GPT-4) \cite{Lu2023} uses GPT-4 as a natural language planner to break down problems into chains of multiple tool combinations (design workflow) and then invoke tools to collaboratively solve problems.
\end{itemize}

\section{Role of Vision Features}

To explore the impact of the proposed method on the LM, the vision features were further removed. Table \ref{B} shows the differences in performance before and after the removal of vision features for different sizes of LMs.

Removing the vision features significantly reduced the performances (approximately 12\% and 9\%, respectively) of both LMs of different sizes, where the performance of the large model reduced relatively slightly. Large performance degradation was mainly generated in the IMG and image-related data. A noticeable performance degradation was also observed in non-image-related data because with the vision features removed, more error CoTs were generated in the rationale generation phase, further disrupting the inference logic in the answer inference phase.

\section{Examples of Case Studies}
\label{C}
\begin{figure*}[t!]
    \centering  
    \centerline{\includegraphics[width=1\textwidth]{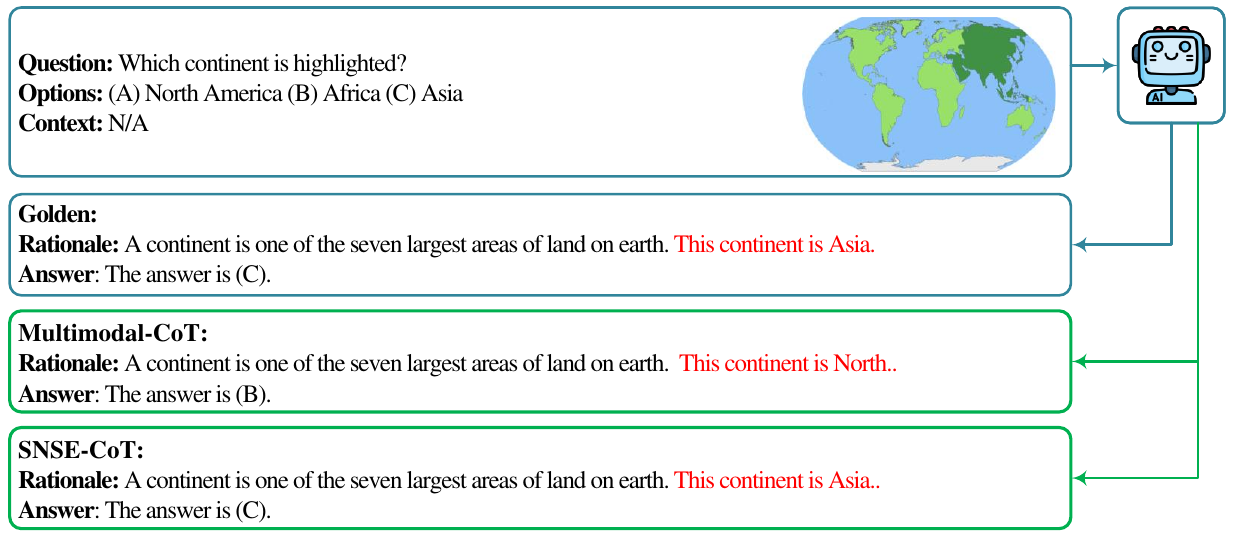}}
    \caption{Example of ID 517.}
    \label{1}
\end{figure*}

\begin{figure*}[t!]
    \centering  
    \centerline{\includegraphics[width=1\textwidth]{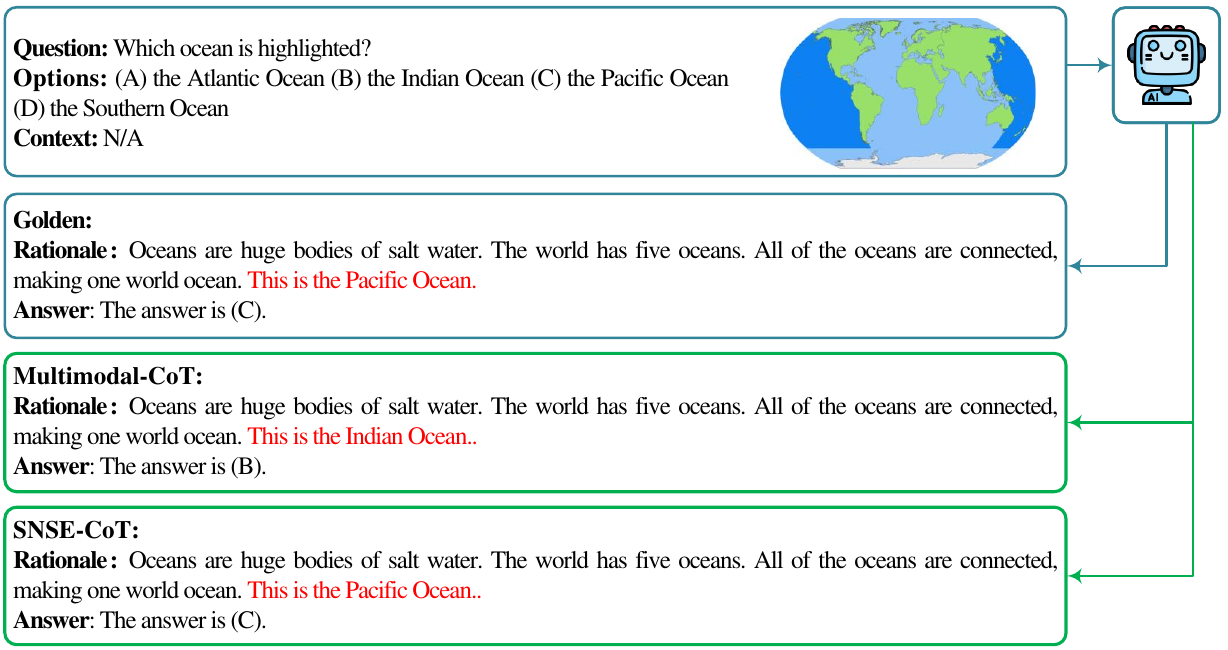}}
    \caption{Example of ID 6902.}
    \label{2}
\end{figure*}

\begin{figure*}[t!]
    \centering  
    \centerline{\includegraphics[width=1\textwidth]{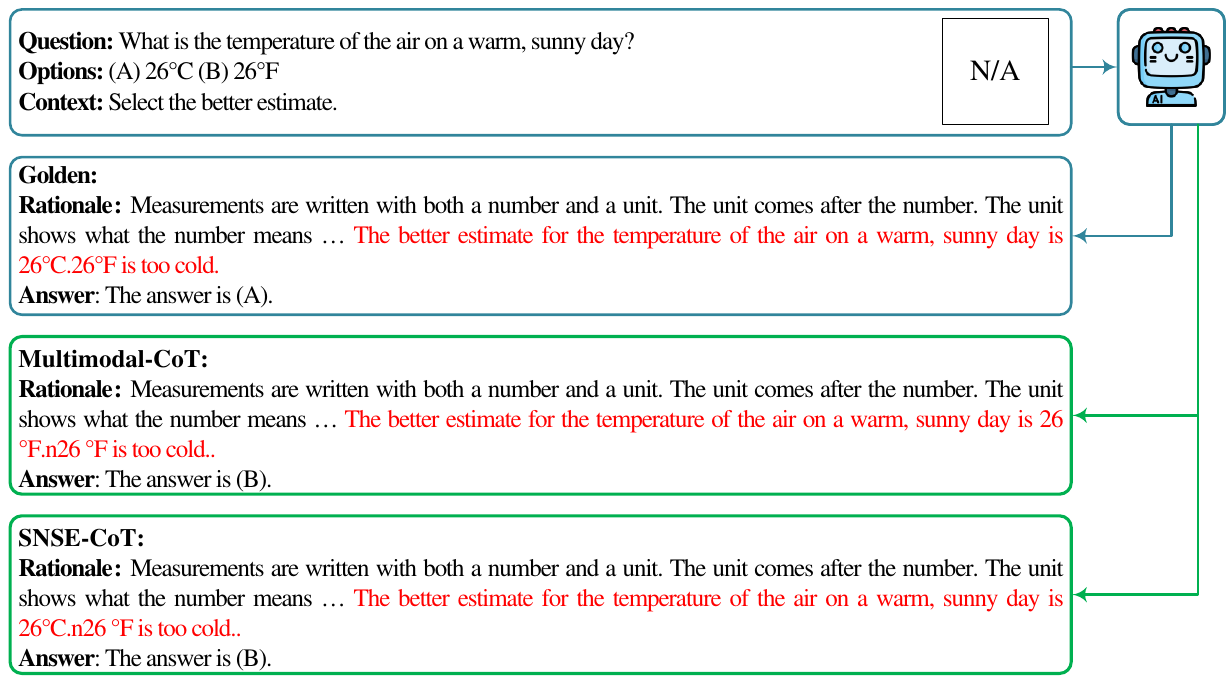}}
    \caption{Example of ID 348.}
    \label{3}
\end{figure*}

\begin{figure*}[t!]
    \centering  
    \centerline{\includegraphics[width=1\textwidth]{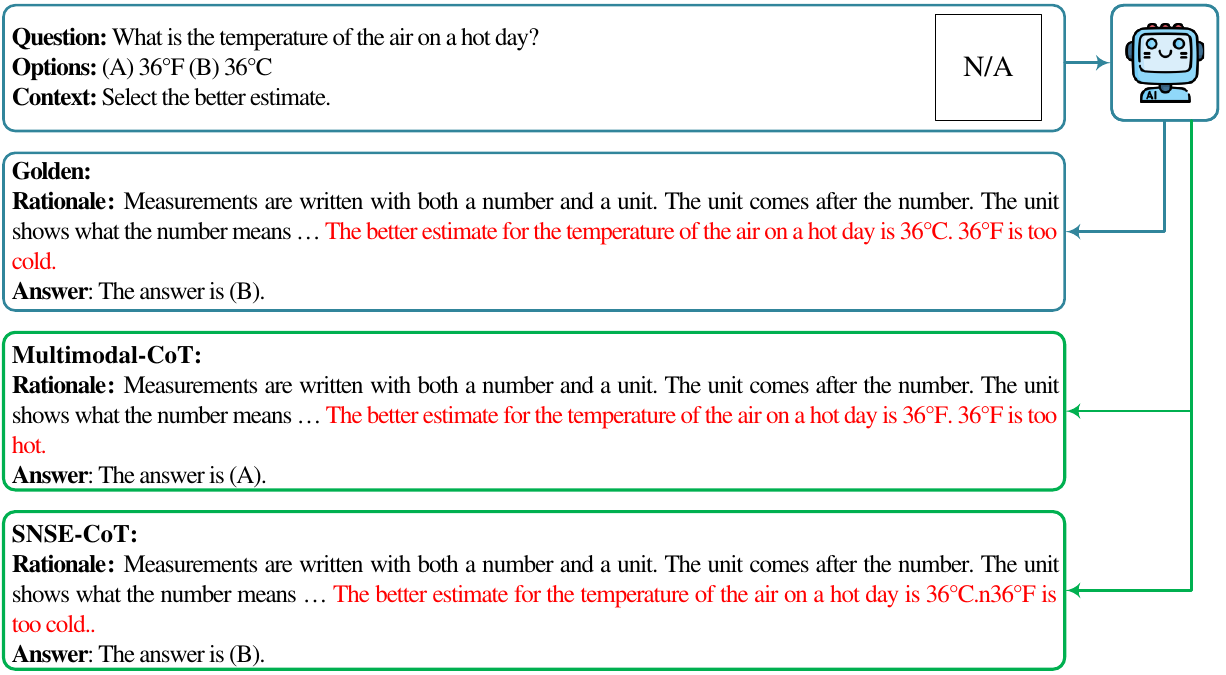}}
    \caption{Example of ID 12785.}
    \label{4}
\end{figure*}

\begin{figure*}[t!]
    \centering  
    \centerline{\includegraphics[width=1\textwidth]{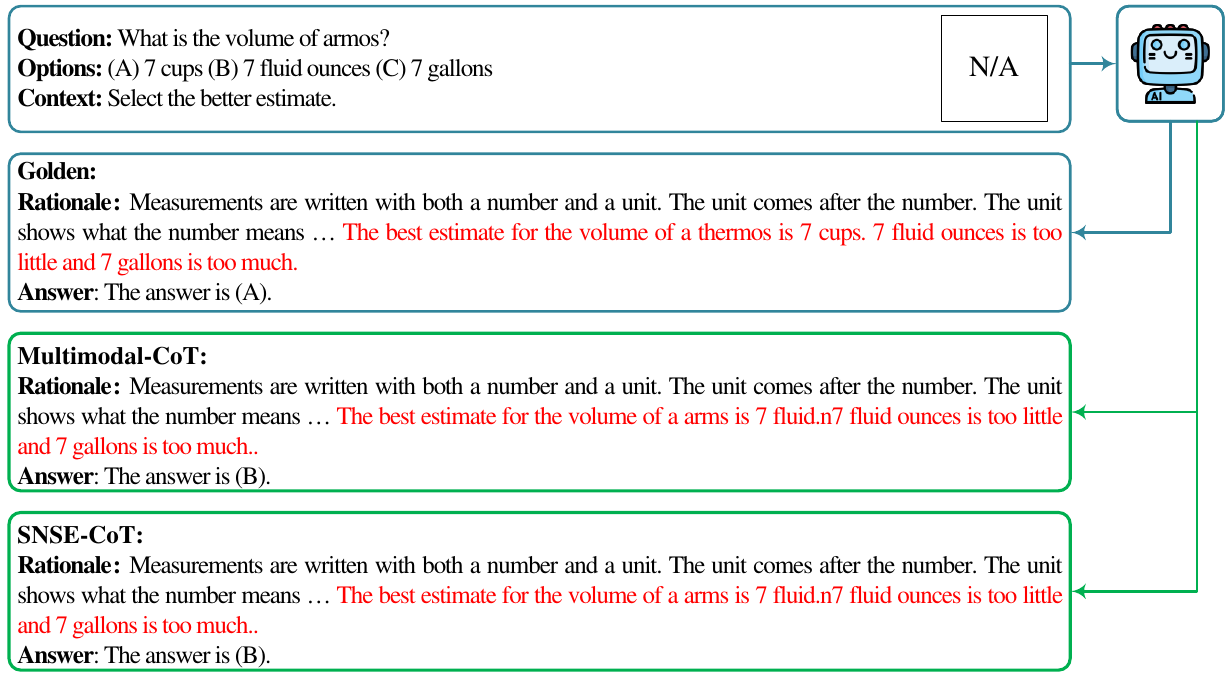}}
    \caption{Example of ID 9653.}
    \label{5}
\end{figure*}

\begin{figure*}[t!]
    \centering  
    \centerline{\includegraphics[width=1\textwidth]{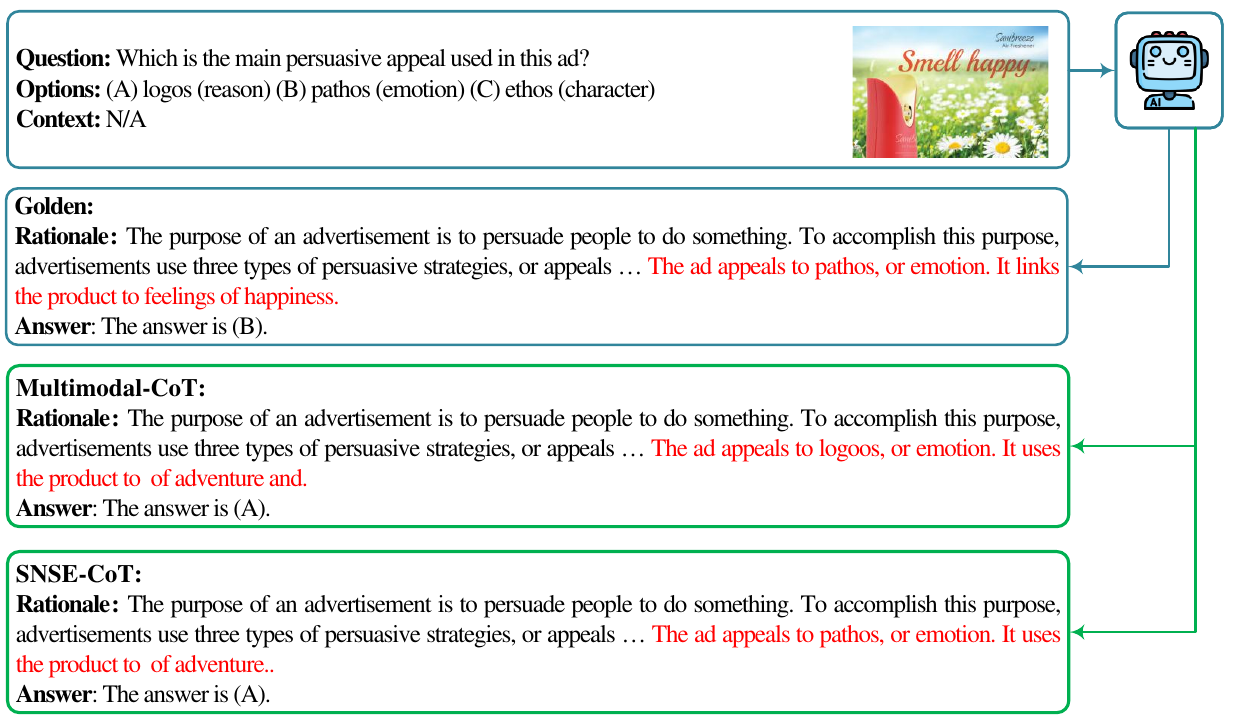}}
    \caption{Example of ID 6964.}
    \label{6}
\end{figure*}

\begin{figure*}[t!]
    \centering  
    \centerline{\includegraphics[width=1\textwidth]{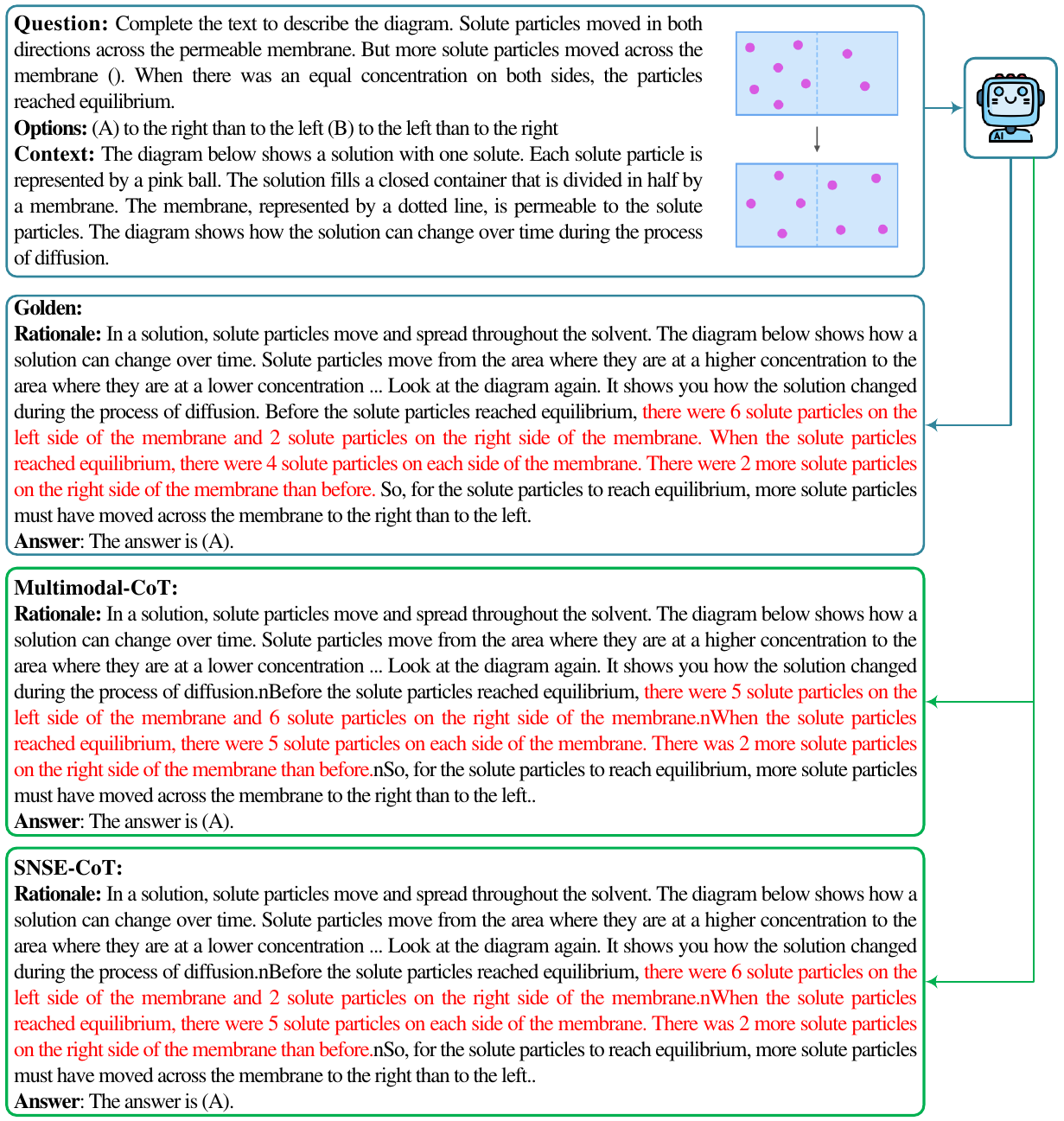}}
    \caption{Example of ID 7301.}
    \label{7}
\end{figure*}

\begin{figure*}[t!]
    \centering  
    \centerline{\includegraphics[width=1\textwidth]{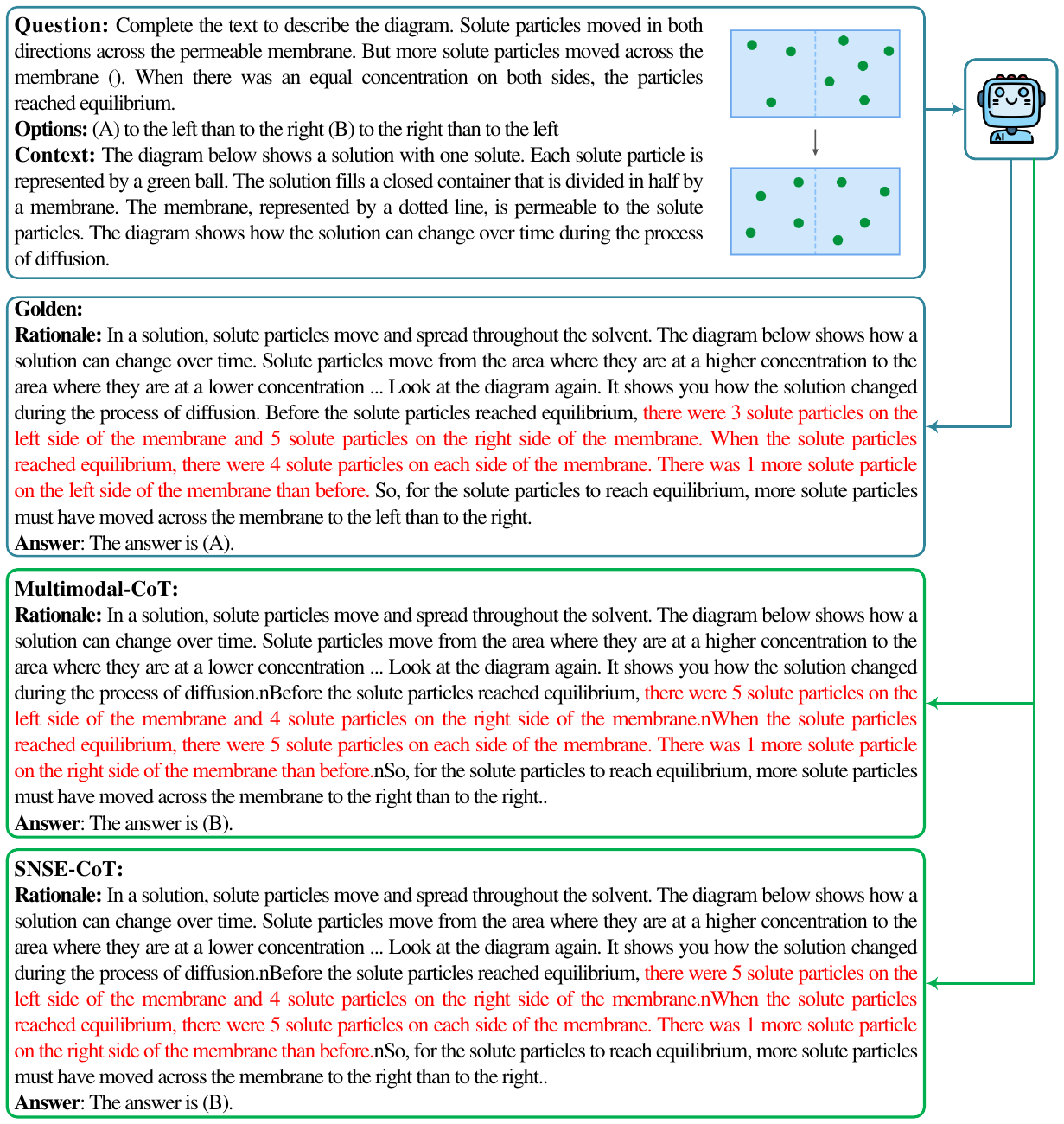}}
    \caption{Example of ID 4740.}
    \label{8}
\end{figure*}

\begin{figure*}[t!]
    \centering  
    \centerline{\includegraphics[width=1\textwidth]{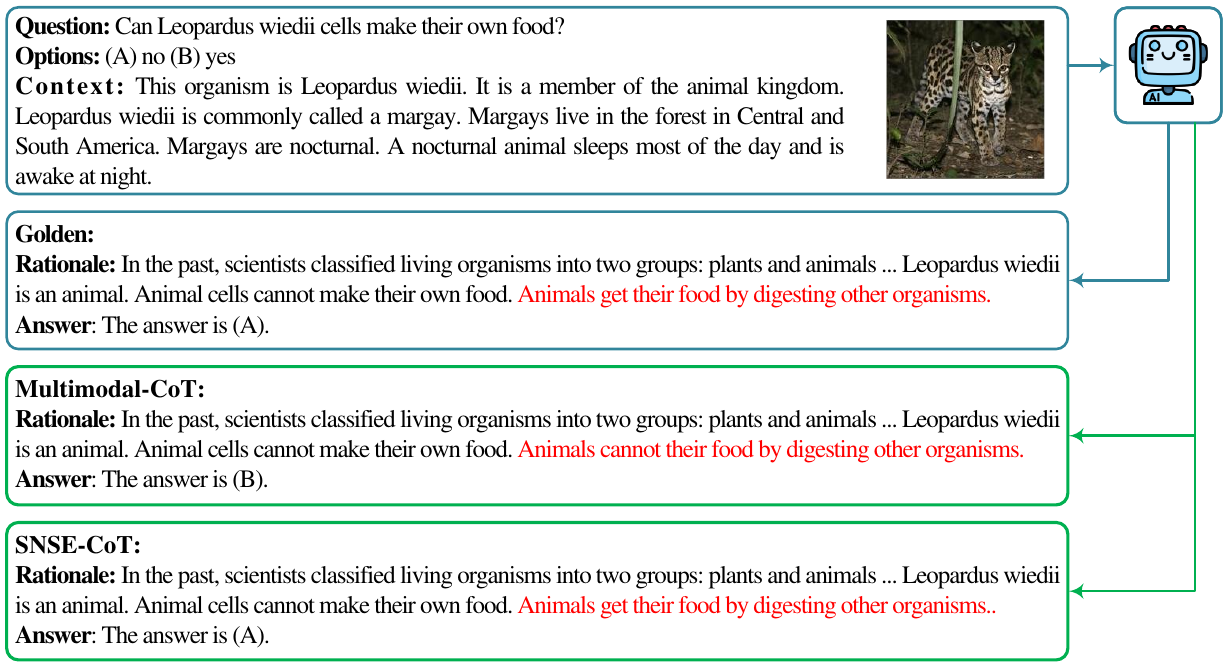}}
    \caption{Example of ID 10130.}
    \label{9}
\end{figure*}

\begin{figure*}[t!]
    \centering  
    \centerline{\includegraphics[width=1\textwidth]{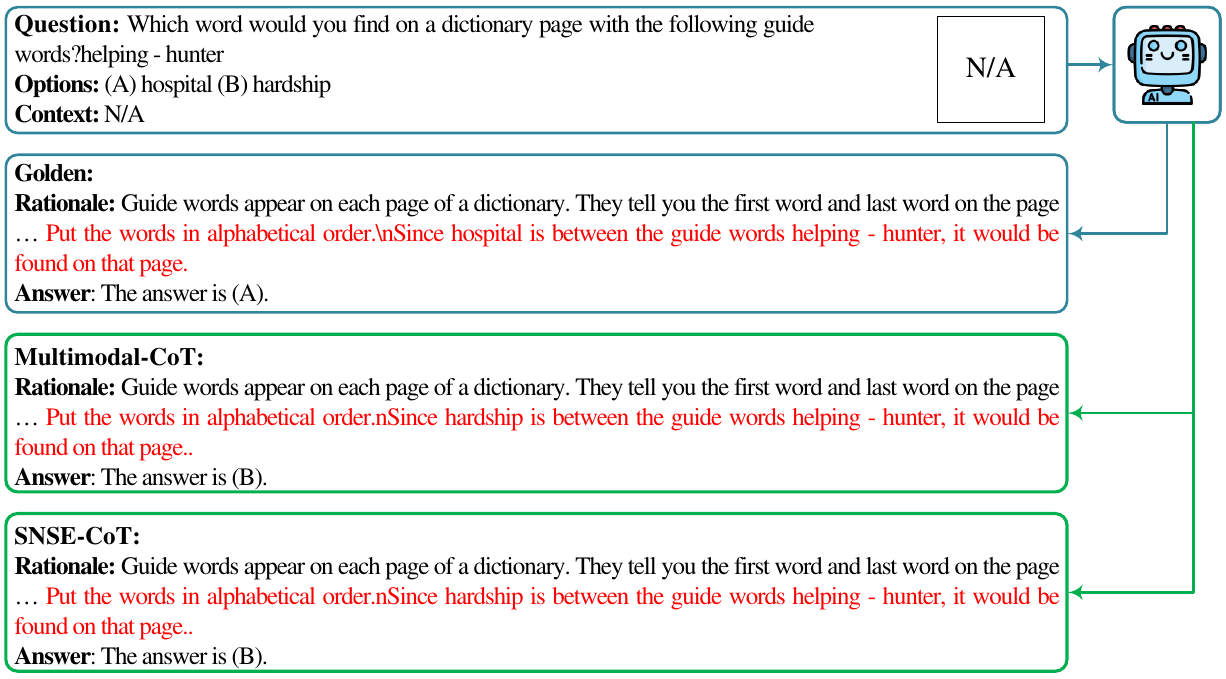}}
    \caption{Example of ID 13068.}
    \label{10}
\end{figure*}

This section presents a detailed case study. As shown in Figures \ref{1} and \ref{2}, the proposed method was well enhanced in terms of identifying the map regions. As shown in Figures \ref{3} and \ref{4}, the unit comparison capability of the model also improved. However, Figures \ref{5} and \ref{6} show that a partially correct CoT may still lead to incorrect answer inferences, and this misdirection is more likely to occur when the conclusion sentence is incorrect. Similarly, as shown in Figures \ref{7} and \ref{8}, guaranteeing the overall correctness of the CoT in terms of number understanding is difficult, even if some numbers can be improved. For common sense problems, models could easily learn common sense that required only memorization (see Figure \ref{9}) but struggled to learn common sense that required logical computation (see Figure \ref{10}).

\end{document}